\documentclass{article}
\usepackage[T1]{fontenc}
\usepackage[utf8]{inputenc}
\usepackage[ngerman]{babel}
\usepackage[skip=4pt]{caption}
\usepackage{booktabs}
\usepackage{dcolumn}
\usepackage{units}
\usepackage{array}
\newcolumntype{x}{>{\scriptsize\raggedright\hspace{0pt}}X}
\makeatother
\usepackage{spconf_new,amsmath,graphicx, amssymb}
\usepackage{graphicx}
\usepackage{xcolor}
\usepackage{float}
\usepackage{multirow}
\usepackage{subcaption}
\usepackage{booktabs}
\usepackage[ruled,linesnumbered]{algorithm2e}

\ninept

\title{DATA-DRIVEN LOW-RANK NEURAL NETWORK COMPRESSION}

\name{Dimitris Papadimitriou$^{\star}$ \qquad Swayambhoo Jain $^{\dagger}$\thanks{Work done at Technicolor AI Lab, Palo Alto, CA. Author emails: \texttt{dimitri@berkeley.edu, swayambhoo.jain@gmail.com} }}
  \address{$^{\star}$ UC Berkeley, $^{\dagger}$ InterDigital AI Lab}

\begin{document}
\maketitle
\begin{abstract}
 Despite many modern applications of Deep Neural Networks (DNNs), the large number of parameters in the hidden layers makes them unattractive for deployment on devices with storage capacity constraints. In this paper we propose a Data-Driven Low-rank (DDLR) method to reduce the number of parameters of pretrained DNNs and expedite inference by imposing low-rank structure on the fully connected layers, while controlling for the overall accuracy and without requiring any retraining. We pose the problem as finding the lowest rank approximation of each fully connected layer with given performance guarantees and relax it to a tractable convex optimization problem. We show that it is possible to significantly reduce the number of parameters in common DNN architectures with only a small reduction in classification accuracy. We compare DDLR with Net-Trim, which is another data-driven DNN compression technique based on sparsity and show that DDLR consistently produces more compressed neural networks while maintaining higher accuracy.
 
\end{abstract}
\section{Introduction}
\label{sec:intro}
Running DNN based applications locally on mobile devices is becoming a necessity for many modern applications. The importance of deploying AI on the edge, in devices such as smartphones, drones and autonomous vehicles, can be mainly attributed to three factors. Using cloud resources to run AI algorithms can lead to delays in inference due to \textit{communication latency}. Furthermore, such communication with the cloud is \textit{energy inefficient} as it requires additional power and is prone to \textit{privacy breaches}, which for instance in the autonomous vehicle industry could have dire consequences.

Recent work suggests that compressing overparametrized DNNs after training leads to a reduction in the overall time and cost of development of DNN based applications \cite{li2020train}. In the context of the ImageNet Large-Scale Visual Recognition Challenge (ILSVRC) \cite{deng2009imagenet} it took multiple years of extensive research and development to reduce the size of initial networks like VGG16 \cite{simonyan2014very} and find computationally efficient alternatives such as MobileNets \cite{howard2017mobilenets}. In comparison, the DNN compression approach can provide an equivalent reduction in a cost-effective way by automating the research for smaller and efficient DNNs.

In this paper, we propose the DDLR approach to compress a given DNN by imposing a low-rank structure on its fully connected layers. While there exist approaches to reduce the number of parameters in a given DNN by imposing structures such as low-rank or sparsity, these approaches are not data-driven and as a consequence they require computationally expensive retraining after parameter reduction \cite{cheng2017survey}. 

Recently, the data-driven sparsity based approach Net-Trim showed that leveraging data during parameter reduction leads to better compression ratios without retraining \cite{aghasi2017net}. While sparsification gives good compression performance for storage and transmission, it is very challenging to get equivalent gains in inference unless special hardware is designed to explicitly exploit the sparsity and custom software implementation is utilized. In contrast, the low-rank based structural approximation that factorizes each parameter matrix as the product of two low dimensional matrices has no such requirements~\cite{wang2019deep}.

Motivated by Net-Trim and the amenability of low-rank structures for faster inference, in this paper we propose a data-driven method that imposes a low-rank structure on the dense layers of a DNN. We formulate this method as a problem of minimizing the rank of the weight matrix of a dense layer under given performance guarantees. This is a non-convex optimization problem which we relax to a tractable convex optimization formulation lying in the family of well known semi-definite programs (SDPs) which can be solved efficiently using off-the-shelf solvers. The proposed DDLR algorithm solves this problem for each dense layer of a pretrained DNN independently, thereby allowing for extensive parallelization of the process.  

Our results show that our method manages to reduce the number of parameters significantly more than Net-Trim while maintaining accuracy levels comparable to the original \textit{uncompressed} network. 
The main advantages of our method can be outlined as follows: \textit{(1)} the imposed low-rank structure allows for large parameter reduction and fast inference via efficient matrix-vector multiplications, \textit{(2)} each layer can be compressed independently allowing for parallel processing, \textit{(3)} the error due to compression in each layer is controlled explicitly and \textit{(4)} our method does not require retraining to achieve high accuracy. 

\section{Related Work}

Given the large size of modern DNN architectures many methods for storage and computational complexity reduction have been proposed in the literature. One commonly used technique for parameter reduction is that of network pruning \cite{liu2018rethinking, blalock2020state}. Network pruning assigns scores to the parameters of a pretrained neural network and removes parameters based on these scores \cite{han2015learning}. A key component of pruning is the need to retrain the model in order to increase accuracy to levels close to the original network. Pruning can be performed on a single parameter basis \cite{aghasi2017net, laurent2020revisiting} or by taking into account groups of parameters that ultimately lead to structured layers amenable for efficient computations \cite{li2016pruning, lin2019towards}. Another branch of parameter reduction techniques is that of low-rank representation of DNN layers. In \cite{tai2015convolutional} the authors present a method to impose a low-rank representation on the convolutional layers of CNNs. Closer to our framework, \cite{sainath2013low} proposes an approach to  impose low-rank structure on the last dense layer of a DNN while training. 

Most of the compression techniques above require further retraining which can be computationally expensive for very large models.  DNN compression without retraining is an important practical problem and recently \cite{aghasi2017net} proposed Net-Trim which leverages data to perform sparsity based parameter reduction and achieve better DNN compression without retraining. However, it is very challenging to extend the compression gains to faster inference speeds as sparsity structure requires custom hardware and software support for that purpose. The off-the-shelf graphical processors (GPUs)
use single instruction, multiple threads execution models, i.e. the same sequence of operations is computed in parallel on different data to accelerate matrix-vector multiplication. The speed of matrix-vector product is then directly linked to the slowest thread, which might be affected by the number of non-zeros in the computation allocation to each thread and the overhead associated with reading the non-zero entries from the chosen compressed sparse storage format. Consequently, it is quite challenging via sparsity structure, since the non-zero entries could be arbitrarily distributed, to get faster matrix-vector products. The low-rank structure on the other hand does not suffer from such issues and provides faster inference using off-the-shelf hardware. Therefore, in this paper we extend Net-Trim \cite{aghasi2017net} to compress DNNs by imposing low-rank structures on the layers.


\section{Method}
In this section we outline our DDLR method that imposes low-rank representations on the weight matrices of the fully connected layers in a pretrained DNN. 
\subsection{DDLR layers}
 Consider a DNN with $L$ dense layers.  Let the $\ell_{\textrm{th}}$ layer of this DNN be a fully connected dense layer with weight matrix $\mathbf{W}_\ell\in\mathbb{R}^{n_{\ell-1}\times n_{\ell}}$ with $n_{\ell-1}$ denoting the dimension of its input and $n_{\ell}$ being the dimension of its output. The corresponding bias of that layer is denoted with $\mathbf{b}_{\ell}\in\mathbb{R}^{n_{\ell}}$. Let also $\mathbf{Y}_{\ell-1}\in\mathbb{R}^{N\times n_{\ell-1}}$ and $\mathbf{Y}_{\ell}\in\mathbb{R}^{N\times n_{\ell}}$ denote the input and output matrices of the $\ell_{\textrm{th}}$ layer respectively, with the number of rows $N$ corresponding to the number of approximation training data in the network input matrix $\mathbf{X} \in \mathbb{R}^{N \times n_0}$. We focus on DNNs that utilize the $\textrm{ReLU}(x) = max(x,0)$ activation function as they form the backbone of DNN architectures. Our method can be generalized to any other activation function that can lead to a convex constraint in~(\ref{initial-optimization}).  Given the input $\mathbf{Y}_{\ell-1}$ the output of layer $\ell$ is obtained as follows
\begin{equation}
\mathbf{Y_{\ell}}= \textrm{ReLU} ( \mathbf{Y}_{\ell-1}\mathbf{W}_{\ell}+\mathbf{1}_{N} \textbf{b}_{\ell}^T), 
\end{equation} 
where $\mathbf{1}_{N}$ is a $N$-dimensional vector of ones. In order to impose a low-rank structure on the weight matrix $\mathbf{W}_{\ell}$ we need to minimize the $rank(\cdot)$ function of that matrix. Given the non-convexity of the $rank(\cdot)$ function we propose solving the following optimization problem
\begin{equation}
\begin{aligned}
& \underset{\mathbf{U}\in\mathbf{R}^{n_{\ell-1}\times n_{\ell}}}{\text{minimize}}
& & ||\mathbf{U}||_* \\
& \text{subject to}
& & ||\textrm{ReLU} ( \mathbf{Y}_{\ell-1}\mathbf{U}+\mathbf{1}_{N} \textbf{b}_{\ell}^T) -\mathbf{Y}_{\ell}||_F\leq \epsilon_{\ell}.
\end{aligned}
\label{initial-optimization}
\end{equation}

The objective function uses the well known nuclear norm relaxation of the $rank(\cdot)$ function in order to obtain a convex objective \cite{fazel2001rank} while the constraint requires the output of the compressed layer to be close to the output of original layer. The layer output error due to compression is controlled by a user specified threshold $\epsilon_{\ell}$. Intuitively, we expect as $\epsilon_{\ell}$ increases the rank of the layer to decrease more since the constraint is becoming more relaxed. However, the constraint in (\ref{initial-optimization}) is non-convex due to the $\textrm{ReLU}$ activation function inside the norm. To alleviate this issue we relax the constraint following~\cite{aghasi2017net}, to obtain the convex constraint
\begin{equation}
\begin{cases}
||(\mathbf{Y}_{\ell-1}\mathbf{U}+\mathbf{1}_{N} \textbf{b}_{\ell}^T-\mathbf{Y}_{\ell})\circ \mathbf{M}_{\ell}||_F^2\leq \epsilon_{\ell}^2\\
(\mathbf{Y}_{\ell-1}\mathbf{U}+\mathbf{1}_{N} \textbf{b}_{\ell}^T)\circ \mathbf{M}_{\ell}'\leq 0
\end{cases},
\label{convex-set}
\end{equation}
where $\mathbf{M}_{\ell}$ is a mask matrix of the same dimension as $\mathbf{Y}_\ell$ selecting the positive entries of $\mathbf{Y}_\ell$ elementwise, i.e. the $(i,j)_{\textrm{th}}$ entry $\textbf{M}_{\ell}^{ij} = 1$ if $ \mathbf{Y}_\ell^{ij} > 0$ and $\mathbf{M}_{\ell}^{ij} = 0$ otherwise. Similarly, $\mathbf{M}_{\ell}'$ is a mask matrix selecting the non-positive entries of $\mathbf{Y}_\ell$ and $\circ$ denotes the matrix Hadamard product. Intuitively, we penalize the deviation of the entries of the compressed layer that correspond to the positive entries of the original layer as the latter are the only ones that are not affected by the ReLU activation function. Furthermore, we allow the entries that correspond to the non-positive entries of the original layer to take any non-positive value. This step allows us to decrease the rank of the weight matrix even more without accumulating extra error. So problem (\ref{initial-optimization}) relaxed using (\ref{convex-set}) can now be written as
\begin{equation}
\begin{aligned}
& \underset{\mathbf{U}\in\mathbf{R}^{n_{\ell-1}\times n_{\ell}}}{\text{minimize}}
& & ||\mathbf{U}||_* \\
& \text{subject to}
& & ||(\mathbf{Y}_{\ell-1}\mathbf{U}+\mathbf{1}_{N} \textbf{b}_{\ell}^T-\mathbf{Y}_{\ell})\circ \mathbf{M}_{\ell}||_F^2\leq \epsilon_{\ell}^2\\
& & &(\mathbf{Y}_{\ell-1}\mathbf{U}+\mathbf{1}_{N} \textbf{b}_{\ell}^T)\circ \mathbf{M}_{\ell}'\leq 0,
\end{aligned}
\label{relaxed-optimization}
\end{equation}
from which we obtain the solution $\hat{\mathbf{U}}_{\ell}$. This formulation allows for imposing structure on the layers while explicitly controlling for the error of the compressed layer output. This problem is a SDP with quadratic constraints and can be solved with most off-the-shelf solvers like SCS~\cite{ocpb:16} and CVXOPT~\cite{andersen2013cvxopt}.

\subsection{Parallel implementation}

The optimization problem~(\ref{relaxed-optimization}) imposes a low-rank structure on a single layer of a network. To compress networks with multiple layers we can compress each layer individually and independently from each other. This process is outlined in Algorithm~\ref{algo-1}, where compression of each layer is an independent of the rest of the layers optimization problem. Given that each layer is compressed independently, the algorithm allows for parallel implementation. The algorithm requires the initial data matrix as input $\mathbf{X}^{N\times n_{0}}$, the original trained weight matrices and biases of the layers and the user specified tolerances $\epsilon_{\ell}$. The output is a sequence of low-rank matrices for each dense layer.

\begin{algorithm}[h]
\SetAlgoLined
	\SetKwInput{Input}{Input~}
	\SetKwInput{Output}{Output~}
	\Input{$\mathbf{X}, \mathbf{W}_{\ell}, \textbf{b}_{\ell}, \;\epsilon_{\ell}, \;\ell = 1,\ldots,L$}
	$\mathbf{Y}_{0}=\mathbf{X}$\\
	\For{$\ell$= 1,\ldots,L}{
		$\mathbf{Y}_{\ell}= \textrm{ReLU}(\mathbf{Y}_{\ell-1}\mathbf{W}_{\ell}+\mathbf{1}_{N} \textbf{b}_{\ell}^T)$
	}
	\For{$\ell$= 1,\ldots, L}{
		$\mathbf{M}_{\ell}^{i,j}=1$ if $\mathbf{Y}_{\ell}^{i,j}>0$,  otherwise $0$ \\
		$\mathbf{M}_{\ell}'^{i,j}=1$ if $\mathbf{Y}_{\ell}^{i,j}\leq0$, otherwise $0$\\
		Solve (\ref{relaxed-optimization})
	}
	\Output{$\hat{\mathbf{U}}_{\ell},\; \ell=1,\ldots, L$}
	\caption{DDLR Algorithm}
	\label{algo-1}
\end{algorithm}

\vspace{0.5cm}
It should be noted that Algorithm~\ref{algo-1} requires the solution of a SDP (line $8$) for each layer. The solution of SDPs can present a computational bottleneck when weight matrices have large dimensions (e.g. the first dense layer of VGG-16) or the number of data samples $N$ used to solve~(\ref{relaxed-optimization}) is large. In such cases, to alleviate these issues one can \textit{1)} solve~(\ref{relaxed-optimization}) to suboptimality using fewer iterations and \textit{2)} use only a subset of the whole training set to solve~(\ref{relaxed-optimization}). Given the size of the layers in the networks studied in the experiments section we will be solving the SDPs to optimality by using only a sample from the original dataset used for training the network.

\section{Experiments}
We demonstrate the effectiveness of our method on three different datasets. An artificial nested spiral, the MNIST~\cite{lecun2010mnist} and the CIFAR-10~\cite{krizhevsky2009learning} datasets. Through experimentation we concluded that compression works well when the number of data samples used to solve~(\ref{relaxed-optimization}) is no less than $5\%$ of the data used to train the network. For this reason, and to deal with the scaling issues of solving SDPs multiple times, we will be training the networks with a subsample of the available data and we will be solving~(\ref{relaxed-optimization}) using a subset of size $N$ of that sample as presented in the following subsections. For each experiment carried out we use two different values of $N$ in order to study the performance of DDLR with respect to that sample complexity. The main method we will be comparing DDLR with is Net-Trim. Net-Trim on which our method is partially based, imposes a sparse structure on the layers post-training by minimizing the induced $\ell_1$ matrix norm~\cite{aghasi2017net}. Both DDLR and Net-Trim can be used to compress DNNs, removing redundancies from the networks and leading to faster inference. We will be comparing the compression level and the resulting accuracies obtained from these two methods. The benefits of DDLR regarding the possible inference speedup was discussed in the related work section. 

For each of the following datasets we utilize Algorithm~\ref{algo-1} for different values of $\epsilon_{\ell}$ to compress a number of the hidden layers. For each experiment we report the relative  accuracy on the test set of the compressed network with respect to the original accuracy of the uncompressed network. We measure the compression by reporting the fraction of parameters needed to be stored for the compressed network with respect to the original. For the sparse matrix experiments we assume that the parameters are stored in COO format. The COO format storage requirement is three times the number of positive entries of a matrix. It should be noted that it is possible to get the same value for the rank for more than one values of $\epsilon_{\ell}$. In such cases, we choose the solution that leads to higher accuracy on the test data. 

\subsection{Spiral dataset}
The first dataset is a spiral of two-dimensional points representing two classes.  The data points were generated by sampling points on the spiral and adding i.i.d. noise uniformly distributed in the interval $[0,3.5]$ for each dimension.
\begin{figure}[h]
	\centering
	\includegraphics[trim=0 0 0 32,clip, scale=0.55]{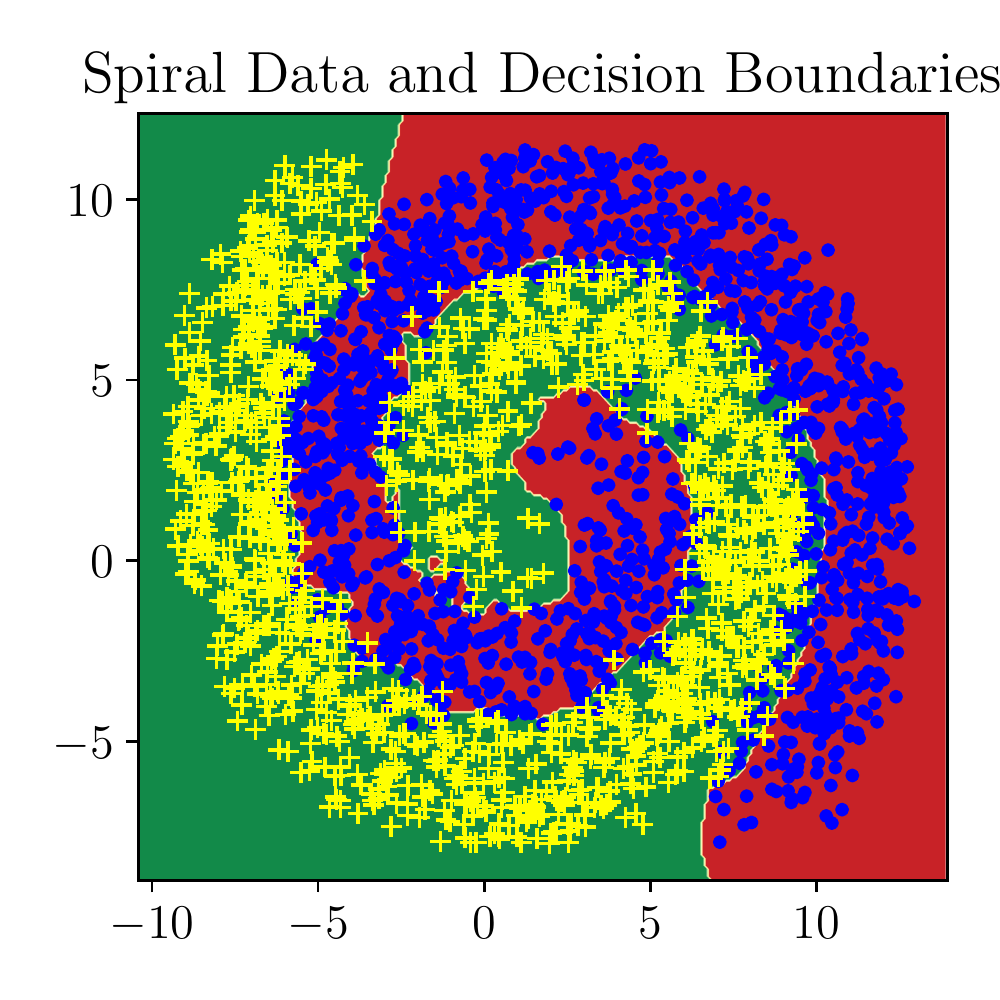}
	\caption{Spiral dataset data and decision boundary.}
	\label{fig:Spiral_New_Data}
\end{figure}
In this experiment, we consider a DNN classifier with two hidden layers to label the points lying on the spiral. The dataset consists of a total of $1024$ points out of which $80\%$ points were used for training the DNN and the remaining $20\%$ were used for testing. The DNN classifier uses the ReLU activation function  except for the last layer where the standard \emph{softmax} function is used. The dimensions of the network layers are $\mathbf{W}_1 \in \mathbb{R}^{2\times 80}, \mathbf{W}_2 \in \mathbb{R}^{80\times 80}$ and $\mathbf{W}_3 \in \mathbb{R}^{80\times 2}$. The DNN was trained by minimizing the cross entropy loss using the stochastic batch gradient descent algorithm with a batch size of $32$ for $1000$ epochs with learning rate $0.001$. The data points along with the decision boundary obtained from the trained DNN are shown in Figure~\ref{fig:Spiral_New_Data}.

We compress only the second layer $\mathbf{W}_2$, as the first and last ones are already low-rank given their dimensions, by utilizing Algorithm~\ref{algo-1}. We implemented the DDLR algorithm under two different scenaria, one using $N=256$ and another using $N=512$ data points for the compression, chosen randomly from the original training dataset. In order to obtain various rank approximations we use the following values for the compression error, $\epsilon_{\ell} \in  [0.02, 0.05, 0.08, 0.1, 0.12, 0.15, 0.2, 0.25, 0.3, 0.5, 0.6]\cdot C$, where $C=\lVert Y_{\ell-1}\rVert_F$ is used for scaling purposes. We use the elbow rule to threshold the singular values of the solution of~(\ref{relaxed-optimization}) to obtain the final low-rank weight matrix. As expected, larger values of $\epsilon_{\ell}$ yield layers with lower rank. For both choices of $N$ DDLR seems to outperform Net-Trim achieving test accuracy close to the original using only $60\%$ of the initial parameters. The original accuracy is recovered with about $80\%$ of the original parameters.

\begin{figure}[h]
	\centering
\hbox{\hspace{-1.4em}    \includegraphics[scale=0.58]{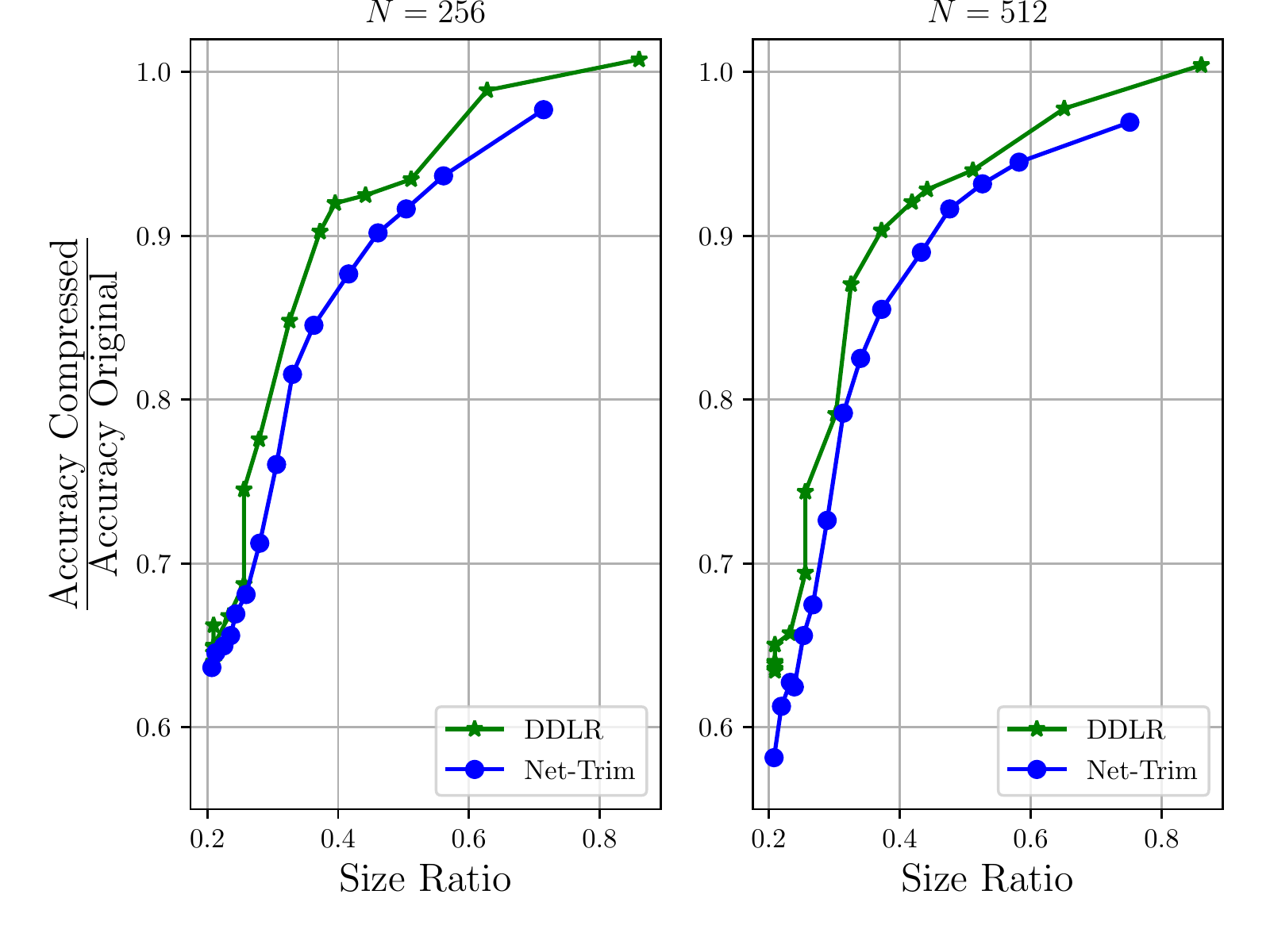}}
	\caption{Relative test accuracy for DDLR and Net-Trim with varying DNN size ratios on Spiral dataset.}
		\label{fig:Spiral_LR_NLR_Comp}
\end{figure}

\subsection{MNIST dataset}

\begin{figure}
	\centering
	\hbox{\hspace{-1.4em}\includegraphics[scale=0.58]{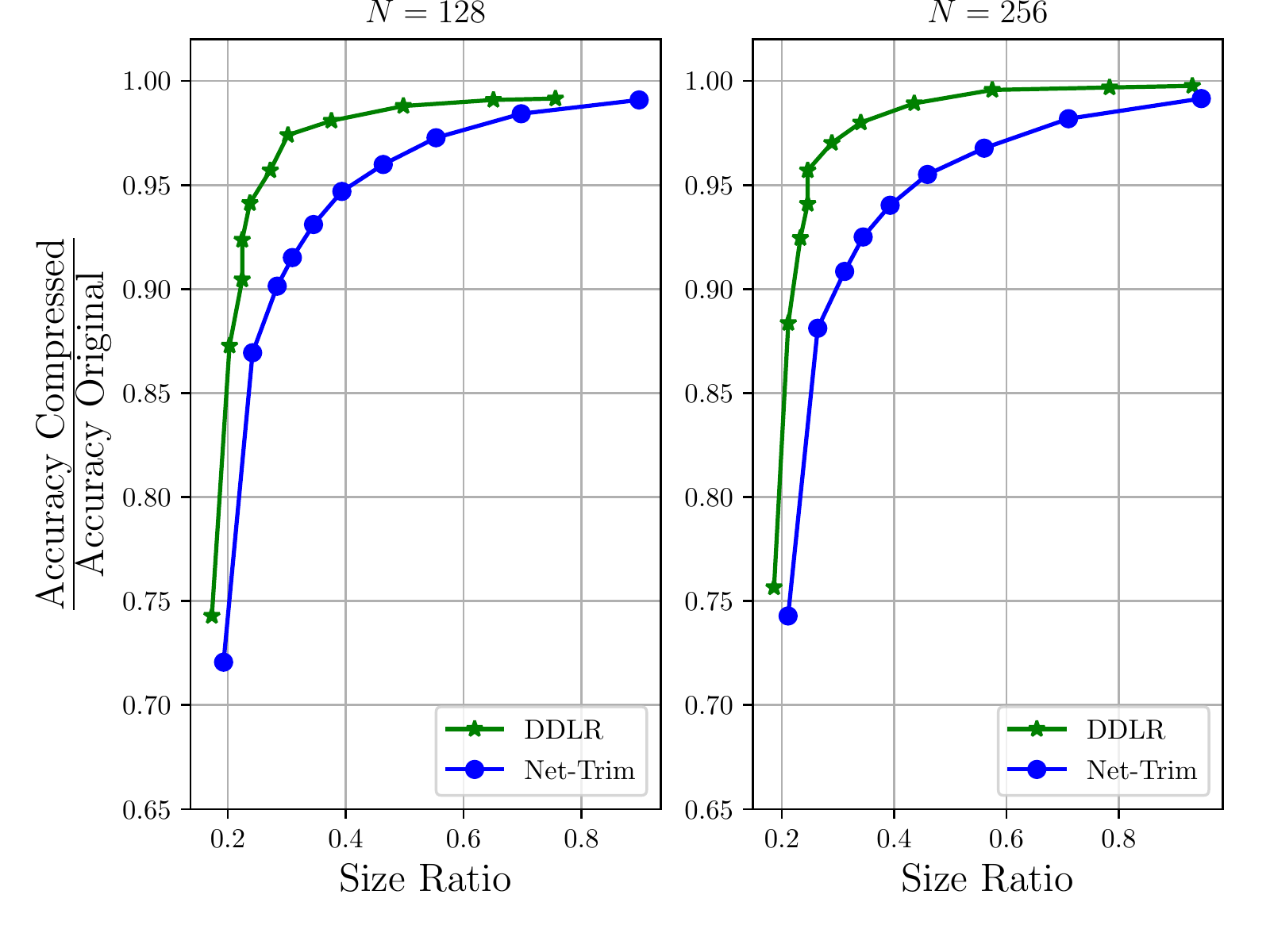}}
	\caption{Relative test accuracy for DDLR and Net-Trim with varying DNN size ratios on MNIST dataset.}
		\label{fig:MNIST_LR_NLR_Comp}
\end{figure}

For the second set of experiments we use the LeNet-5 CNN architecture~\cite{lecun1998gradient} to classify the handwritten digits of the MNIST dataset. LeNet-5 has three dense layers of dimensions $\mathbf{W}_1\in\mathbb{R}^{256\times120},\mathbf{W}_2\in\mathbb{R}^{120\times84}$ and $\mathbf{W}_3\in\mathbb{R}^{84\times10}$ that follow the convolutional layers. We train LeNet-5 using $1024$ data samples from the original dataset, a $80\%$-$20\%$ train-test split, a batch size of $64$, $30$ epochs and a learning rate of $0.001$. Using Algorithm~\ref{algo-1} we impose a low-rank structure on the first two dense layers of LeNet-5 $\mathbf{W_1}$ and $\mathbf{W}_2$ using the following values of $\epsilon_{\ell}=[0.01,0.02,0.04,0.06,0.08,0.1,0.12,0.14,0.16,0.2,0.3]\cdot C$ for each layer, with $C=\lVert\mathbf{Y}_{\ell-1}\rVert_F$ being a scaling constant. We use $N=128$ and $N=256$ number of samples out  of the $1024$ data points to compress the network. Figure~\ref{fig:MNIST_LR_NLR_Comp} presents the relative accuracy for different size ratios. DDLR achieves high compression while maintaining sufficient accuracy. With a $70\%$ reduction in the number of parameters DDLR can achieve a test accuracy less than $4\%$ lower than that of the original network for both values of $N$ while for $N=256$ with a $40\%$ reduction in parameters the accuracy is almost identical to the original. Interestingly, we observe that even for $N=128$ samples, which corresponds to slightly more than $10\%$ of the original data, we are able to obtain high compression associated with high accuracy.

\subsection{CIFAR-10 dataset}

\begin{figure}
	\centering
	\hbox{\hspace{-1.4em}\includegraphics[scale=0.58]{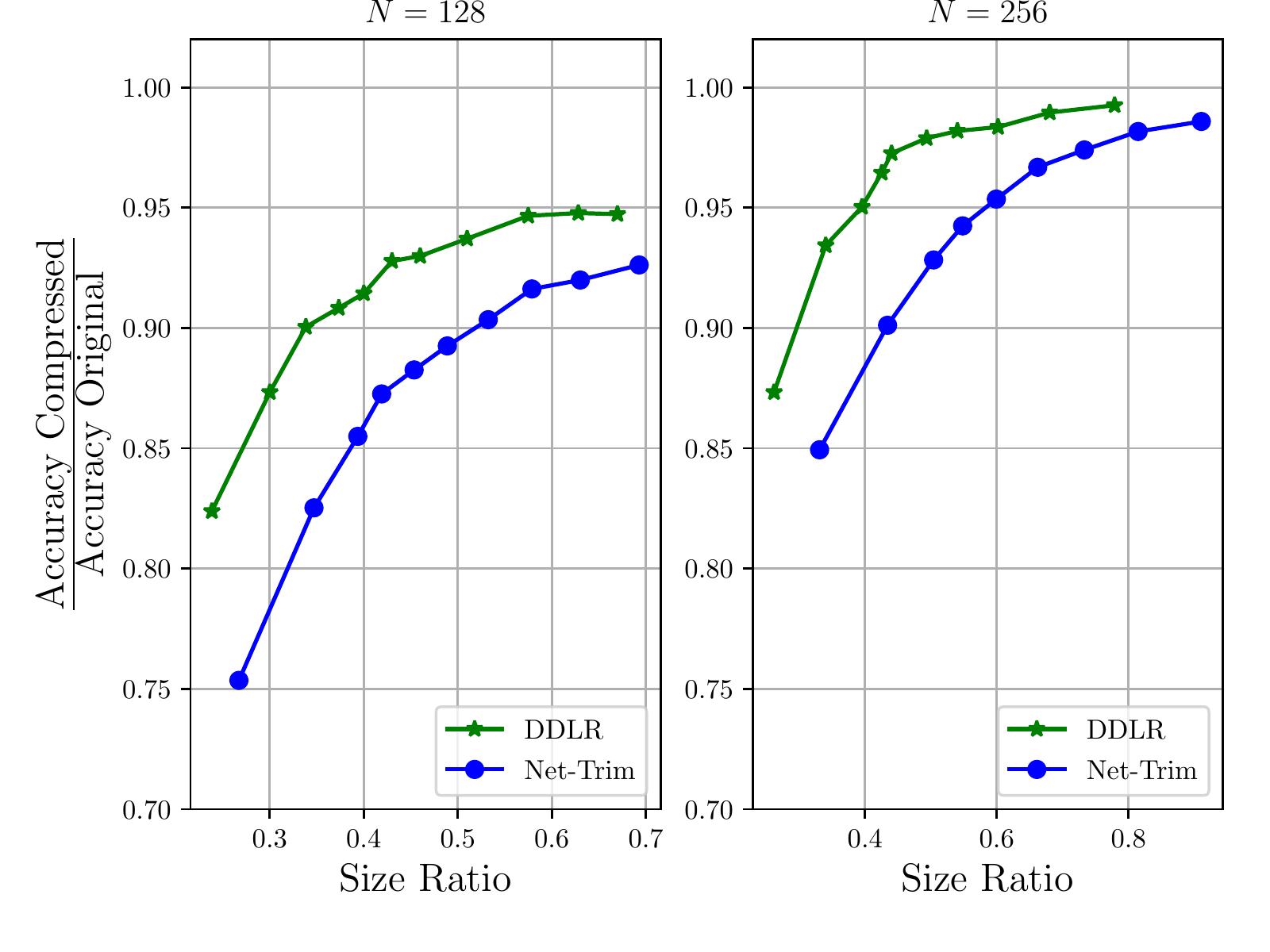}}
	\caption{Relative test accuracy for DDLR and Net-Trim with varying DNN size ratios on CIFAR-10 dataset.}
		\label{fig:CIFAR_LR_NLR_Comp}
\end{figure}

For the final set of experiments we classify the CIFAR-10 image dataset  using again the LeNet-5 network. For CIFAR-10 the dense layers of Lenet-5 have dimensions $\mathbf{W}_1\in\mathbb{R}^{400\times120},\mathbf{W}_2\in\mathbb{R}^{120\times84}$ and $\mathbf{W}_3\in\mathbb{R}^{84\times10}$. We use $2048$ data points from CIFAR-10 to train our network and $N=128$ and $N=256$ subsamples for compression. For this experiment we compress the first two dense layers $\mathbf{W}_1$ and $\mathbf{W}_2$ using the following values for $\epsilon_{\ell}=[0.01,0.02,0.04,0.06,0.08,0.1,0.12,0.14,0.16,0.2,0.3]\cdot C$, where $C=\lVert\mathbf{Y}_{\ell-1}\rVert_F$. As expected, for larger $N$ both methods perform better with DDLR still outperforming Net-Trim. Quite astonishingly, we observe that with only $50\%$ of the original parameters DDLR achieves an accuracy less than $2\%$ worse in comparison to the accuracy of the original network for $N=256$. For the same relative accuracy on the other hand Net-Trim reduces only by $20\%$ the total number of parameters needed to be stored.  Figure~\ref{fig:CIFAR_LR_NLR_Comp} contains the curves of the relative accuracies with respect to the parameter ratio.

\section{conclusion and future work}
The DDLR Algorithm is an end-to-end approach that compresses a pretrained DNN by imposing low-rank structures on the fully connected layers while controlling for the overall accuracy decrease in the compressed DNN. We demonstrate in a number of datasets and DNN architectures that high parameter reduction can be achieved at a small loss in accuracy while requiring no retraining. Such reduction can be very significant for storing already trained models on edge devices. Furthermore, low-rank structured layers allow for fast matrix-vector multiplications without the need for specialized hardware which reduce inference time, something that is of vital importance for AI applications, especially on the edge.

The results of the experiments presented in this paper are rather encouraging but also limited due to the computational bottleneck of solving large scale SDPs. This drawback poses an interesting problem that requires theoretical and algorithmic development in future research. An interesting approach in that direction is a reformulation of the optimization problem~(\ref{relaxed-optimization}) in order to be solved using the Alternating Direction Method of Multipliers (ADMM) \cite{boyd2011distributed}. Such an approach can provide better scalability that will allow for compression of networks with larger hidden layers that have been trained on large datasets. Another interesting direction is to understand the impact of quantization on DNNs already compressed using DDLR, as such quantization can lead to significant additional reduction in the size of the networks. 


\bibliographystyle{IEEEbib}
\bibliography{references}

\end{document}